\documentclass[10pt,twocolumn]{article}

\usepackage{cvpr}
\usepackage{times}
\usepackage{amsmath}
\usepackage{amssymb}
\usepackage{multirow}
\usepackage{array}
\usepackage{subfig}

\makeatletter
\@namedef{ver@everyshi.sty}{}
\makeatother
\usepackage{tikz}
\usepackage{tikzscale}
\usepackage{pgfplots}
\usepackage{arydshln}

\usepackage{fixltx2e}

\usepackage[font={small,it}]{caption}

\usepackage[pagebackref=true,breaklinks=true,letterpaper=true,colorlinks,bookmarks=false]{hyperref}
\usepackage{pgf}
\cvprfinalcopy 

\def\cvprPaperID{5026} 

\newcolumntype{C}[1]{>{\centering\let\newline\\\arraybackslash\hspace{0pt}}m{#1}}

\ifcvprfinal\fi

\graphicspath{ {./figures/}
				{./figures/method/}
				{./figures/implementation/}
				{./figures/results/}}

\begin{document}

\title{A Push-Pull Layer Improves Robustness of Convolutional Neural Networks}

\author{Nicola Strisciuglio\\
University of Groningen\\
\\
{\tt\small n.strisciuglio@rug.nl}
\and
Manuel Lopez-Antequera\\
University of Groningen\\
University of Malaga\\
{\tt\small m.lopez.antequera@rug.nl}
\and
Nicolai Petkov\\
University of Groningen\\
\\
{\tt\small n.petkov@rug.nl}
}

\maketitle

\begin{abstract}
We propose a new layer in Convolutional Neural Networks (CNNs) to increase their robustness to several types of noise perturbations of the input images. 
We call this a push-pull layer and compute its response as the combination of two half-wave rectified convolutions, with kernels of opposite polarity. It is based on a biologically-motivated non-linear model of certain neurons in the visual system that exhibit a response suppression phenomenon, known as push-pull inhibition. 
   
We validate our method by substituting the first convolutional layer of the LeNet-5 and WideResNet architectures with our push-pull layer.
We train the networks on non-perturbed training images from the MNIST, CIFAR-10 and CIFAR-100 data sets, and test on images perturbed by noise that is unseen by the training process. We demonstrate that our push-pull layers contribute to a considerable improvement in robustness of classification of images perturbed by noise, while maintaining state-of-the-art performance on the original image classification task.
\end{abstract}

\section{Introduction}
Convolutional Neural Networks (CNNs) are routinely used in many problems of image processing and computer vision, such as large-scale image classification~\cite{alexnet}, semantic segmentation~\cite{segnet}, optical flow~\cite{opticalflow}, stereo matching~\cite{stereo}, among others.
They became a de~facto standard in computer vision and are gaining increasing research interest.
The success of CNNs is attributable to their ability of learning representations of input training data in a hierarchical way, which yelds state-of-the-art results in a wide range of tasks.
The availability of appropriate hardware, namely GPUs and deep learning dedicated architectures, to facilitate huge amounts of required computations has favoured their spread, use and improvement. 

A number of breakthroughs in image classification were achieved by end-to-end training of deeper and deeper architectures. AlexNet~\cite{alexnet}, VGGNet~\cite{vggnet} and GoogleNet~\cite{googlenet}, which were composed of eight, 19 and 22 layers, respectively, pushed forward the state-of-the-art results on large-scale image classification.
Subsequently, learning of extremely deep networks was made possible with ResNet~\cite{resnet}, whose architecture based on stacked bottleneck layers and residual blocks helped alleviate the problem of vanishing gradients.
Such very deep networks, with hundreds or even a thousand layers, contributed to push the classification accuracy even higher on many benchmark data sets for image classification and object detection.
With WideResNet~\cite{wideresnet}, it was shown that shallower but wider networks can achieve better classification results without increasing the number of learned parameters.

These networks suffer from reliability problems due to output instability~\cite{stabletraining}, i.e. small changes of the input cause big changes of the output.
Some approaches to increase the stability of deep neural networks to noisy images make use of data augmentation, i.e. new training images are created by adding noise to the original ones.
This approach, however, improves robustness only to those classes of perturbation of the images represented by the augmented training data and requires that this robustness is learned: it is not intrinsic to the network architecture.
In~\cite{stabletraining}, a more structured solution to the problem was proposed, where a loss function that controls the optimization of robustness against noisy images was introduced. 

Instead, we use prior knowledge about the visual system to guide the design of a new component for CNN architectures: in this paper, we propose a new layer called \emph{push-pull layer}.
We were inspired by the push-pull inhibition phenomenon that is exhibited by some neurons in area V1 of the visual system of the brain~\cite{Taylor595}.
Such neurons are tuned to detect specific visual stimuli, but respond to such stimuli also when they are heavily corrupted by noise. 
The inclusion of this layer contributes to an increase in robustness of CNNs to noise and contrast changes of the input images, \emph{while maintaining state-of-the-art performance on the original classification task}.
This comes without an increase in the number of parameters and with a negligible increase in computation.

\noindent Our contributions are summarized as follows:

\begin{itemize}
\item We propose a new biologically-inspired layer for CNN architectures. It implements the push-pull inhibition mechanism that is exhibited by some neurons in the visual system of the brain, which respond to the stimuli they are tuned for also when they are corrupted by noise. 
\item We validate our method by including the proposed push-pull layer into state-of-the-art residual network architectures and training them from scratch on the task of image classification on several datasets. We study the effect of using the proposed push-pull layer in the first layer of CNN architectures. Our push-pull layer intrinsically embues the network with improved robustness to noise without increasing the model size.
\item We show the impact of the proposed method by comparing the performance of networks with and without the push-pull layer on the problem of classification of noisy images. Our proposal improves accuracy on noisy images while maintaining performance on the original images.
\item We provide an implementation of the proposed push-pull layer as a new layer for CNNs in PyTorch. 
\end{itemize}

\section{Related works}
\paragraph{Data augmentation. }
The success of CNNs and deep learning in general can be attributed to the unparalleld representation capacity of these models, enabled by their size and hierarchical nature.
However, this large capacity can become problematic as it can be hard to avoid overfitting to the training set. Early work achieving success on large scale image classification~\cite{alexnet} noted this and included data augmentation schemes, where training samples were modified by means of transformations of the input image that do not modify the label, such as rotations, scaling, cropping, and so on~\cite{alexnet}.
Data augmentation schemes can also be used to allow the network to learn invariances to other transformations not present on the training set but that can be expected to appear when deploying the network.



The main drawback of data augmentation is that the networks acquire robustness only to the classes of perturbations used for training~\cite{stabletraining}.
Additionally, these invariances are learned, whereas some invariances could instead be directly introduced as part of the architecture.

\paragraph{Prior knowledge in deep architectures.}

Domain specific knowledge can be used to guide the design of deep neural network architectures. In this way, they better represent the problem to be learned in order to increase efficiency or performance. For example, Convolutional Neural Networks are a subset of general neural networks that encode translational invariance.

Specific architectures or modules have been designed to encode properties of other problems.
%
For instance, steerable CNNs include layers of steerable filters to compute orientation-equivariant feature response maps~\cite{steerablecnn}. 
They achieve  rotational equivariance by computing the responses of feature maps at a given set of orientations.
In Harmonic CNNs, rotation equivariance was achieved by substituting convolutional filters with circular harmonics~\cite{hnet}.
In~\cite{sphericalnet}, a formulation of spherical cross-correlation was proposed, enabling the design of Spherical CNNs, suitable for application on spherical images.

\paragraph{Biologically-inspired models. }
One of the first biologically inspired models for Computer Vision was the neocognitron network~\cite{Fukushima1980}. The architecture consisted of layers of S-cells and C-cells, which were models of simple and complex cells in the visual system of the brain. The network was trained \emph{without a teacher}, in a self-organizing fashion.
As a result of the training process, the neocognitron network had a structure similar to the hierarchical model of the visual system formalized by Hubel and Wiesel~\cite{Hubel1962}. 

The filters learned in the first layer of a CNN trained on natural images resemble Gabor kernels and the receptive fields of neurons in area V1 of the visual system of the brain~\cite{Marcelja80}. 
This strenghtens the connection between CNN models and the visual system. However, the convolutions used in CNNs are linear operations, and are not able to correctly model some non-linear properties of neurons in the visual system, e.g. cross orientation suppression and response saturation.
These properties were achieved by a non-linear model of simple cells in area V1, named CORF (Combination of Receptive Fields), used in image processing for contour detection~\cite{AzzopardiPetkov2012} or for delineation of elongated structures~\cite{Azzopardi2015}.

A layer of non-linear convolutions inside CNNs was proposed in~\cite{volterracnn}. The authors were inspired by neuro-physiological studies of non-linear processes in early stages of the visual system, and modeled them by means of Volterra convolutions.

\section{Push-pull networks}

We propose a new layer that can be used in existing CNN architectures to improve their robustness to different classes of noises. We call it \emph{push-pull} layer as its design  is inspired by the structure and functions of some  neurons in area V1 of the visual system of the brain that exhibit a phenomenon known as \emph{push-pull} inhibition~\cite{Kremkow2016}. Such neurons have excitatory and inhibitory components that respond to stimuli of opposite polarity.
Their responses are combined in such a way that these neurons strongly respond to specific visual stimuli, also when they are corrupted by noise.
We provide a wider discussion about the biological inspiration of the proposed push-pull layer in the Appendix~\ref{appendix1}. In the rest of the Section, we explain the details of the proposed layer.

\subsection{Push-pull layer}

\begin{figure}[!t]
	\begin{center}
		\input{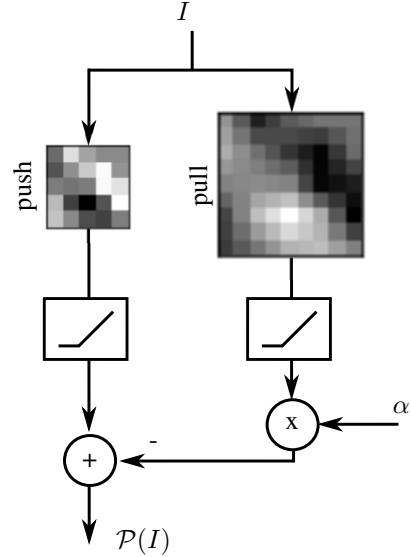}
	\end{center}
	\caption{Architectural scheme of the push-pull layer.}
	\label{fig:pushpull}
\end{figure}

We design the push-pull layer $\mathcal{P}(I)$ using two convolutional kernels, which we call \emph{push} and \emph{pull} kernels, that model the \emph{excitatory} and the \emph{inhibitory} components of the push-pull neuron, respectively. 
The pull kernel typically has a larger support region than the push kernel and its weights are computed by inverting and upsampling the push kernel~\cite{Taylor595}.
We implement \emph{push-pull} inhibition by subtracting a fraction $\alpha$ of the response of the pull component from the one of the push component. We model the activation functions of the push and pull receptive fields by using non-linearities after the computation of the push and pull response maps.
In Figure~\ref{fig:pushpull} we show an architectural sketch of the proposed layer.

We define the response of a push-pull layer as:
\begin{equation}
\nonumber
\mathcal{P}(I) = \Theta\left(k \ast I\right) - \alpha \Theta(-k_{\uparrow h} \ast I) 
\end{equation}

\noindent where $\Theta(\cdot)$ is a rectifier linear unit (ReLU) function, $\alpha$ is a weighting factor for the response of the pull component which we term inhibition strength. Finally, $\uparrow h$ indicates upsampling by a scale factor $h$.

In Figure~\ref{fig:examples} we show the response maps of a push convolutional kernel only (second row) in comparison with those of a push-pull layer (third row), computed on input samples corrupted with increasing levels of Gaussian noise (first row).
One can observe how the push-pull layer is able to detect the features, which were learned in the training phase, more reliably than the push only component, even when the input is corrupted by high levels of noise. This effect is determined by the pull component, which suppresses the responses of the push convolutional kernel due to noisy and spurious patterns.

\subsection{Use of the push-pull layer}
\begin{figure}[!t]
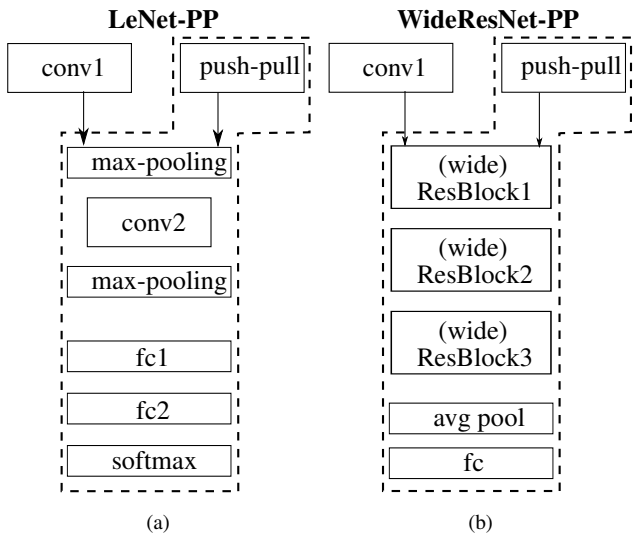

	\centering
\vspace{-5mm}
	\subfloat[] {
		\input{figures/implementation/lenetPP-paper.eps_tex}
		\label{fig:conf1}
	}
	\subfloat[]{
		\input{figures/implementation/wideresnetPP-paper.eps_tex}
		\label{fig:conf2}
	}
\caption{Modified (a) LeNet and (b) WideResNet architectures. We substitute the first layer of convolutions (conv1) with our push-pull layer. The suffix `PP' in the network names stands for push-pull. The new architectures are highlighted by the dashed lines. }
\label{fig:implementation}
\end{figure}

We implemented a push-pull layer for CNNs in PyTorch
and deploy it by substituting the first convolutional layer of existing CNN architectures.
In Figure~\ref{fig:implementation}a-b, we show sketches of modified versions of LeNet-5 and WideResNet, respectively. We replaced the first convolutional layer \emph{conv1} with our \emph{push-pull} layer. The resulting architecture is surrounded by the dashed contour line. In the rest of the paper we utilize the suffix `-PP' to indicate that the concerned network model has a push-pull layer as the first layer.

\begin{figure*}[!t]
\small
\renewcommand{\arraystretch}{1}
\begin{tabular}{c|c} 
\multicolumn{2}{c}{\textbf{Comparison of convolution and push-pull layer response maps on noisy input}} \\ \hline 
~ & Input images with different levels of noise \\
 	& \includegraphics[width=145mm ]{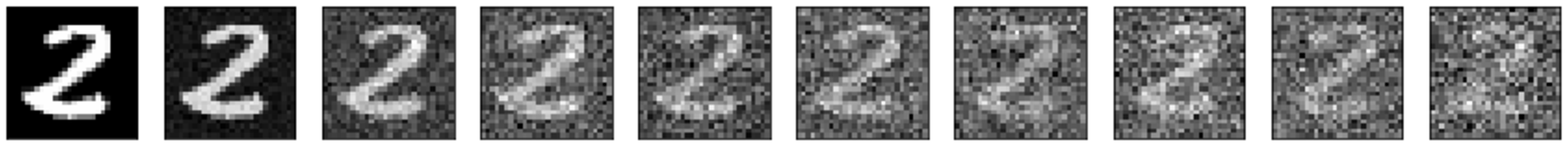} \\ \hline 
\textit{conv kernel} & Response maps of the convolutional kernel \\
  \includegraphics[height=12mm]{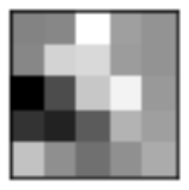} & \includegraphics[width=145mm]{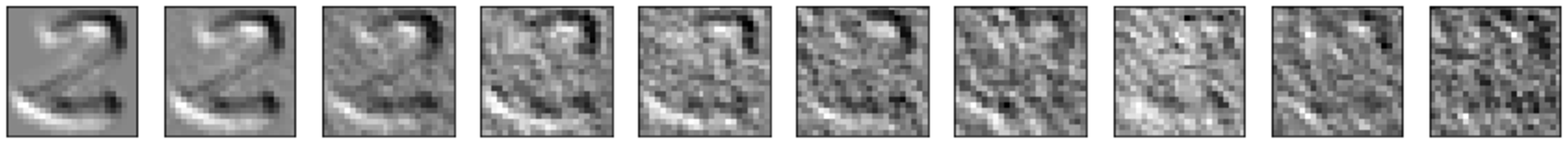}\\ \hdashline
\textit{push-pull layer} & Response maps of the push-pull layer \\   
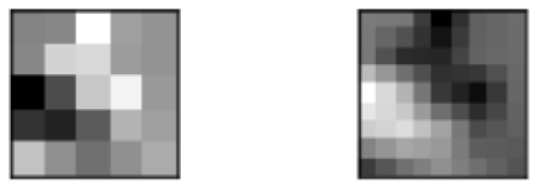 & \includegraphics[width=145mm]{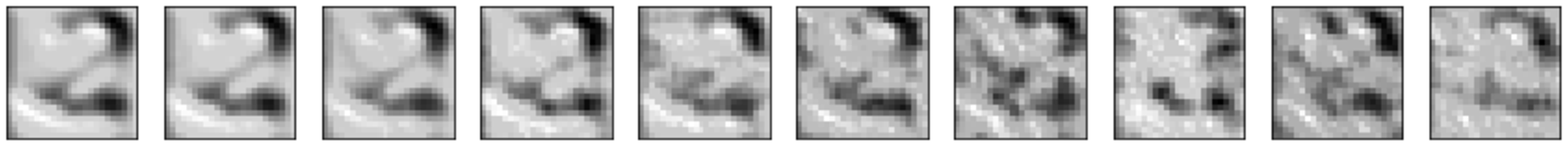}\\ \hline
\end{tabular}
\vspace{2mm}
\caption{Images of a digit perturbed with increasing level of added Gaussian noise (first row). The response maps of a convolutional kernel in the second row show instability with respect to perturbed inputs. Our push-pull layer is more robust to noise as shown in the response maps in the third row.
}
\label{fig:examples}
\end{figure*}

In this work, we train the modified architectures from scratch. One can also replace the first layer of convolutions of an already trained model with our push-pull layer. In such case, however, the model requires a fine-tuning procedure so that the layers succeeding the push-pull layer can adapt to the new response maps, as the responses of the push-pull layer are different from those of the convolutional layers (see the second and third rows in Figure~\ref{fig:examples}). 

In principle, the push-pull layer may be used at any depth level in deep network architectures as a substitute of convolutional layers. However, its implementation is related to the behavior of neurons in early stages of the visual system of the brain, where low-level processing of the visual stimuli is performed.

\section{Experiments}
We carried out extensive experiments to validate the effectiveness of our push-pull layer for improving the robustness of existing networks to perturbations of the input image.
We include the push-pull block in the LeNet-5 and WideResNet architectures, and train several configurations of such networks on the training sets of the MNIST and CIFAR data sets, respectively.

We test the performance of the networks on test images perturbed by Gaussian and speckle noise with increasing variance. Furthermore, we  change the contrast of the images in the test set and apply Poisson noise, and study the effects that it has on the classification accuracy.
We compare the results obtained by CNNs with and without push-pull layers. The results that we report were obtained by replacing the first convolutional layer with a push-pull layer with upsampling factor $h=2$ and inhibition strenght $\alpha=1$. In Section~\ref{sec:sensitivity}, we study the sensitivity of the classification performance with respect to different configurations of the push-pull layer.

\subsection{LeNet-5 and MNIST}
\label{sec:lenet}
The MNIST data set is composed of 60k images of handwritten digits (of size $28 \times 28$ pixels), divided into 50k training and 10k test images. The data set has been widely used in computer vision to benchmark algorithms for object classification. LeNet-5  is one of the first convolutional networks~\cite{lenet}, and is composed of two convolutional layers for feature extraction and three fully connected layers for classification. It achieved remarkable results on the MNIST data set, and is considered one of the milestones of the development of CNNs. We use it in the experiments for the simplicity of its architecture, which allows to better understand the effect of the push-pull layer on the robustness of the network to noise.

We configured different LeNet-5 models by changing the number of convolutional filters in the first and second layer (note that the size of the fully connected layer changes accordingly to the number of filters in the second convolutional layer). We implemented push-pull  versions of {LeNet-5} by substituting the first convolutional layer with our push-pull layer. In Table~\ref{tab:lenetconf}, we report details on the configuration of the LeNet-5 models. The letter `P' in the model names indicate the use of the proposed push-pull layer.

\begin{table}[!t]
\centering
\small
\renewcommand{\arraystretch}{1.1}
\begin{tabular}{c|cc|C{18mm}} 
\hline 
~ & \multicolumn{2}{c|}{\bfseries conv-net} & ~ \\ \cline{2-3}
 \bfseries model name &  \bfseries 1st layer &  \bfseries 2nd layer & \bfseries fc-net \\ \hline \hline
 A & $6$ (c)  & $16$ (c) & $128,64,10$\\
 B & $6$ (c) &  $8$ (c) & $64,32,10$ \\
 C & $4$ (c) & $16$ (c) & $128,64,10$ \\
 D & $4$ (c) & $8$ (c) & $64,32,10$ \\ \hdashline 
 PA & $6$ (pp)  & $16$ (c) & $128,64,10$ \\
 PB & $6$ (pp) &  $8$ (c) & $64,32,10$ \\
 PC & $4$ (pp) & $16$ (c) & $128,64,10$ \\
 PD & $4$ (pp) & $8$ (c) & $64,32,10$ \\ \hline \hline
\end{tabular}
\caption{Configurations of the LeNet-5 architecture used in the experiments on the MNIST data set. The label (c) indicate a convolutional layer, while (pp) indicates a push-pull layer.}
\label{tab:lenetconf}
\end{table}

\begin{figure*}[!t]
	\footnotesize
	\begin{center}
		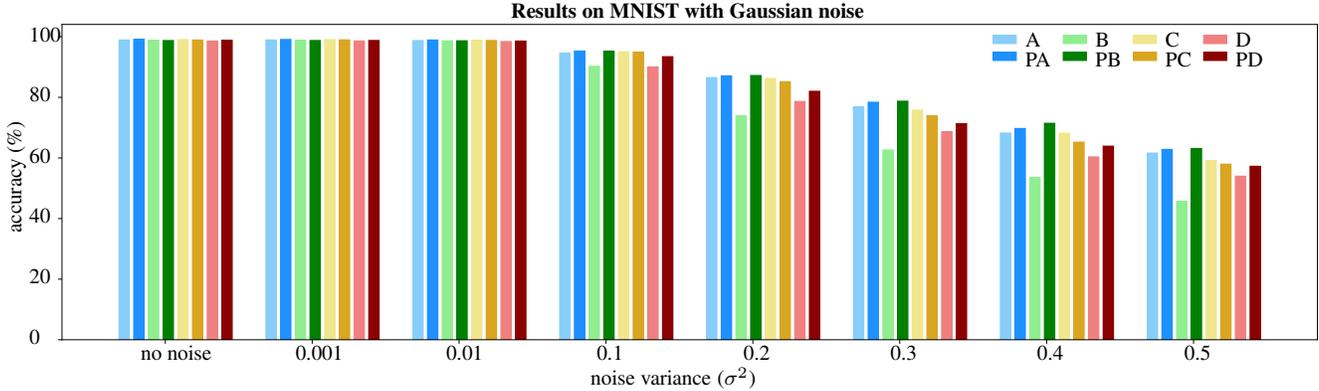
	\end{center}
	\vspace{-6mm}
	\caption{Results of LeNet-5 (lighter colors - A, B, C, D) and push-pull LeNet-5 (darker colors - PA, PB, PC, PD) on the MNIST data set with Gaussian noise.}
	\label{fig:lenet-gaussian}
\end{figure*}

We report the results achieved on the MNIST data set perturbed with Gaussian noise of increasing variance in Figure~\ref{fig:lenet-gaussian}. When the variance of the  noise increases above $\sigma^2=0.1$, the improvement of performance determined by the use of the push-pull layer is noticeable ($A=86.5\%$, $PA=87.1\%$ - $B=73.91\%$, $PB=87.2\%$ - $C=86.2\%$, $PC=85.14\%$ - $D=78.62\%$, $PD=82\%$, for Gaussian noise with $\sigma^2=0.2$), witnessing an increase of the generalization capabilities of the networks and of their robustness to noise. Only in the case of the model C, the push-pull layer does not provide a beneficial effect on the classification results. We report results on test images with speckle noise in the supplementary materials.

In Figure~\ref{fig:lenet-poisson}, we compare the results achieved by the different LeNet-5 models with the push-pull layer (darker colors - PA, PB, PC, PD) with those of the original LeNet-5 (lighter colors - A, B, C, D) on the MNIST test set images perturbed by change of image contrast and addition of Poisson noise. We use different factors $C$ to increase or decrease the contrast of the input images $I$, and produce new images $I_C = (I-0.5) * C + 0.5$.

The LeNet-5 models with our push-pull layer considerably outperform their convolution-only counterparts when the contrast of noisy test images decreases and the images are corrupted by Poisson noise. It is interesting that the convolutional models A and D show a considerable drop of classification performance when the contrast level is  lower than $C=0.5$. We conjecture that this is probably due to specialization of the networks on the characteristics of the images in the training set. The models B and C achieve more stable results when the contrast level  is 0.5 and 0.4, but their performance decrease considerably when the Poisson noise is applied on images with lower contrast. In all cases, the push-pull versions of LeNet-5 show higher robustness to noise perturbations of the images than their respective convolutional versions. 
It is worth pointing out that the classification accuracy on the original data is not affected by the use of the push-pull layer ($A=98.93\%$, $PA=99.1\%$ - $B=98.85\%$, $PB=98.78\%$ - $C=99.06\%$, $PC=98.91\%$ - $D=98.58\%$, $PD=98.84\%$).

\begin{figure}[!t]
\vspace{-8mm}
	\footnotesize
	\begin{center}
		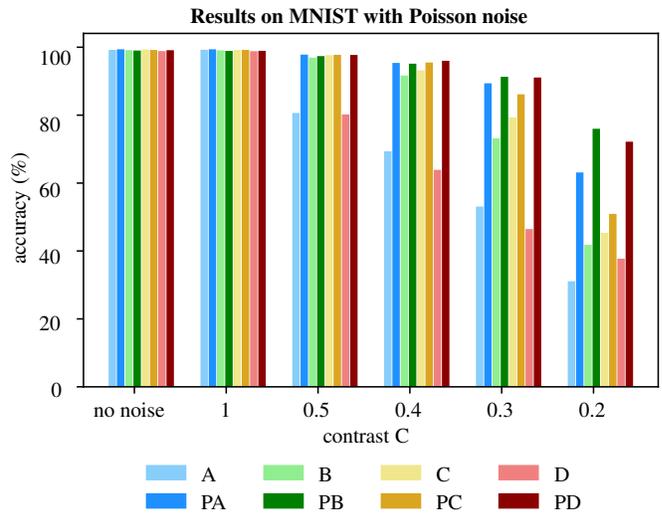
	\end{center}
	\vspace{-6mm}
	\caption{Results of LeNet-5 (lighter colors) and push-pull LeNet-5 (darker colors) on MNIST with Poisson noise.}
	\label{fig:lenet-poisson}
\end{figure}

\begin{table}[!t]
\centering
\small
\renewcommand{\arraystretch}{1.1}
\begin{tabular}{cc|c|cc|cc}
\multicolumn{7}{c}{\bfseries Results achieved with WRN-28-10} \\ \hline
\multicolumn{3}{r|}{\bfseries data set} & \multicolumn{2}{c}{CIFAR-10} & \multicolumn{2}{|c}{CIFAR-100} \\ \hline 
\multicolumn{3}{r|}{\bfseries push-pull layer} & no & yes & no & yes \\ \hline \hline
\multirow{8}{*}{\rotatebox{90}{gaussian noise ($\sigma^2$)}} & 0 & \multirow{8}{*}{\rotatebox{90}{accuracy (\%)}} & \bfseries 96.1 & 95.93 & \bfseries 80.82 & 80.39 \\
~ & 0.0001 & ~ & \bfseries 95.33 & \bfseries 95.33 & 77.76 & \bfseries 78.13 \\
~ & 0.0005 & ~ & 90.94 & \bfseries 92.38 & 67.78 & \bfseries 69.05 \\
~ & 0.001 & ~ & 85.52 & \bfseries 88.94 & 56.81 & \bfseries 59.77 \\
~ & 0.005 & ~ & 48.56 & \bfseries 63.15 & 18.2 & \bfseries 24.1 \\
~ & 0.01 & ~ & 29.68 & \bfseries 43.61 & 8.12 & \bfseries 11.33 \\
~ & 0.02 & ~ & 18.58 & \bfseries 25.86 & \bfseries 3.91 & 3.62 \\
~ & 0.03 & ~ & 14.55 & \bfseries 18.32 & \bfseries 2.68 & 1.77 \\ \hline 

\multirow{8}{*}{\rotatebox{90}{speckle noise ($\sigma^2$)}} & 0 & \multirow{8}{*}{\rotatebox{90}{accuracy (\%)}} & \bfseries 96.1 & 95.93 & \bfseries 80.82 & 80.39 \\
~ & 0.0001 & ~ & \bfseries 95.88 & 95.45 & \bfseries 79.85 & 79.24 \\
~ & 0.0005 & ~ & \bfseries 94.85 &  94.44 & \bfseries 77.2 &  76.35 \\
~ & 0.001 & ~ & 93.52 & \bfseries 93.56 & \bfseries 73.77 & \bfseries 73.77 \\
~ & 0.005 & ~ & 83.73 & \bfseries 86.65 & 53.97 & \bfseries 56.57 \\
~ & 0.01 & ~ & 73.6 & \bfseries 79.28 & 39.23 & \bfseries 43.99 \\
~ & 0.02 & ~ & 59.95 & \bfseries 69.29 & 22.78 &\bfseries 28.92 \\
~ & 0.03 & ~ & 51.69 & \bfseries 62.74 & 15.63 & \bfseries 20.47 \\ \hline

\multirow{6}{*}{\rotatebox{90}{poisson noise ($C$)}} & 0.4 & \multirow{6}{*}{\rotatebox{90}{accuracy (\%)}} &   20.15 & \bfseries 31.34 & 3.28 & \bfseries 5.71 \\
~ & 0.5 & ~ & 28.31 &  \bfseries 42.84 & 7.04 &  \bfseries 11.09 \\
~ & 1 & ~ &  \bfseries 96.1 & 95.93 & \bfseries 80.82 & 80.39 \\
~ & 1.5 & ~ & 65.17 & \bfseries 66.05 & 15.39 & \bfseries 18.8 \\
~ & 2 & ~ & 62.56 & \bfseries 70.6 & 13.55 & \bfseries 16.02 \\
~ & 2.5 & ~ & 53.61 & \bfseries 65.27 & 9.02 &\bfseries 11.2  \\  \hline  \hline

\end{tabular}
\caption{Results achieved on noisy CIFAR-10 and CIFAR-100 test sets by WideResNet with 28 layers and widen factor equal to 10 (WRN-28-10), with and without the push-pull layer. `yes' and `no' labels indicate wheter the results reported in the corresponding columns are obtained by the network with or without the push-pull layer, respectively.}
\label{tab:resultsCIFAR}
\end{table}

\subsection{WideResNet and CIFAR}
CIFAR-10 and CIFAR-100~\cite{cifar} are data sets of natural images (of size $32 \times 32$ pixels) organized in 10 and 100 classes, respectively. The data sets contain 60k images, divided into 50k for training and 10k for test, and are widely used to benchmark the performance of state-of-the-art CNN architectures for image classification. 

We carried out experiments on the CIFAR data sets using several configurations of the WideResNet CNN~\cite{wideresnet}.
WideResNet is based on the popular residual network ResNet and implements a mechanism to widen the size of the network (i.e. the number of residual blocks per layer). The authors showed that widening the network contributes to improvement of classification results and generalization better than increasing its depth (i.e. the number of layers). We train several WideResNet models with and without push-pull layers, and test their performance on the CIFAR test set images, which we perturbed with noise and changes of contrast. In the rest of the section, we call WRN-L-W a WideResNet model with L layers and a widening factor equal to W. For Example, WRN-16-10 is a WideResNet with 16 layers and a widening factor of 10. In the case $W=1$, the architecture is reduced to a ResNet model~\cite{resnet}.

%
%

\begin{figure}[!t]
	\footnotesize
	\begin{center}
		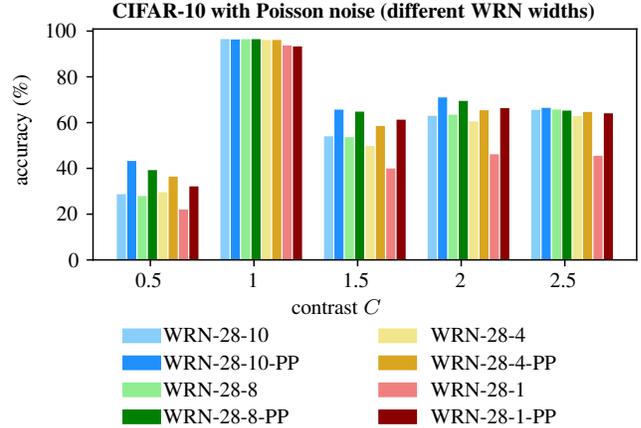
	\end{center}
	\vspace{-6mm}
	\caption{Results achieved on CIFAR-10 with Poisson noise by WRN architectures with different widen factors. The Poisson noise is applied after changing the contrast $C$ of the original images.}
	\label{fig:wrn-poisson-wider}
\end{figure}
\begin{figure}[!t]
	\footnotesize
	\begin{center}
		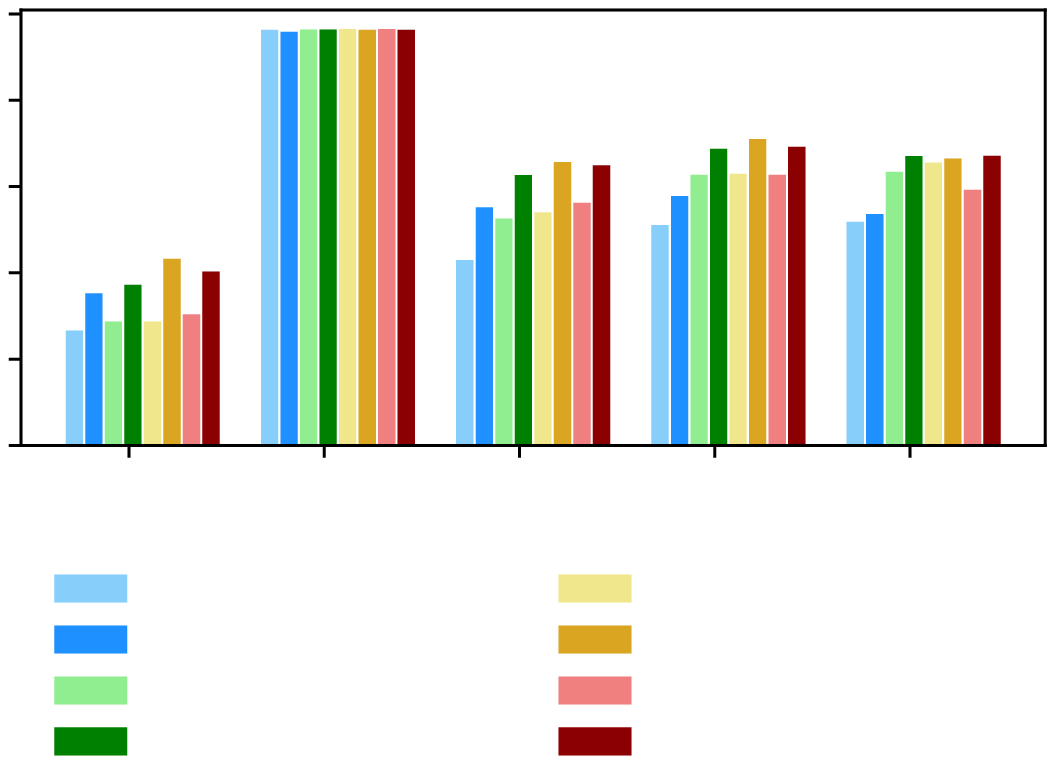
	\end{center}
	\vspace{-6mm}
	\caption{Results achieved on CIFAR-10 with Poisson noise by WRN architectures with different depths. The Poisson noise is applied after changing the contrast $C$ of the original images.}
	\label{fig:wrn-poisson-depth}
\end{figure}

\begin{figure*}[!t]
	\footnotesize
	\begin{center}
		\input{figures/results/cifar10-widen-paper.eps_tex}
	\end{center}
\vspace{-6mm}
	\caption{Results achieved on CIFAR-10 with Gaussian noise by WideResNet with different widen factors. The light color bars report the results of WideResNet without the push-pull layer, while darker color bars the results of networks with push-pull layer (-PP suffix).}
	\label{fig:widenWRN}
\end{figure*}

\begin{figure*}[!t]
	\footnotesize
	\begin{center}
		\input{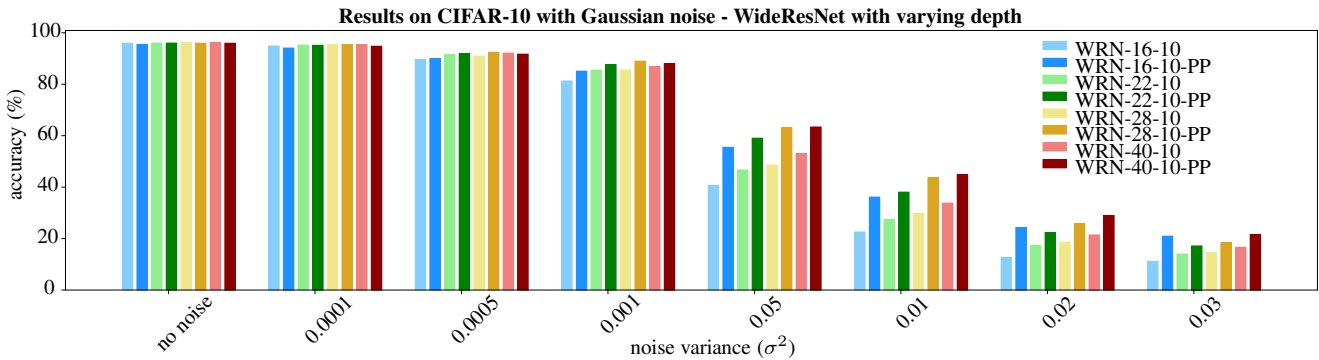}
	\end{center}
\vspace{-6mm}
	\caption{Results achieved on CIFAR-10 with Gaussian noise  by WideResNet with different depths. The light color bars report the results of WideResNet without the push-pull layer, while darker color bars the results of networks with push-pull layer (-PP suffix).}
	\label{fig:depthWRN}
\end{figure*}

In Table~\ref{tab:resultsCIFAR}, we report the results that  we achieved on the CIFAR-10 and CIFAR-100 data sets, using a WideResNet with 28 layers and a widening factor of 10, with and without the push-pull layer. We considered WRN-28-10 as it is reported to achieve very high classification performance on CIFAR data sets~\cite{wideresnet}. We test on the original CIFAR test sets and also on noisy versions of them which we created by adding Gaussian, speckle and Poisson noise. We add Poisson noise after changing the contrast of the original image  as explained in Section~\ref{sec:lenet}. We observed a substantial improvement of the robustness to noise perturbations of the input images by models that deploy the proposed push-pull layer as a substitute of the first convolutional layer. The classification accuracy is in some case up to $15\%$ higher than that of networks without push-pull layers. A point of strenght of using the push-pull layer is that the number of parameters of the modified models does not increase with respect to the original architecture. The weights of the push kernel are learned by backpropagation, similarly to learning the weights of convolutional kernels, while the weights of the pull kernels are derived from those of the push kernels by upsampling and inversion of polarity.

We studied the influence of the push-pull layer on the performance of several WideResNet models, which we configured varying the widening factor and the depth. 
We reports the results  that we achieved on the CIFAR-10 test set images perturbed by Poisson noise and contrast changess, using WideResNet models with different widen factor in Figure~\ref{fig:wrn-poisson-wider} and different depths in Figure~\ref{fig:wrn-poisson-depth}.
In Figure~\ref{fig:widenWRN} and~\ref{fig:depthWRN}, we show the results of similar experiment performed on the CIFAR-10 test set with added Gaussian noise by using WideResNet models with different widen factors and depths, respectively. In all cases, the push-pull layer contributes to an improvement of robustness to noise perturbations of the test images. We also report results on images with speckle noise in the supplementary materials.
It is worth to highlight the case of the WRN-28-1-PP network, which has widen factor of 1 and corresponds to a ResNet model with a push-pull layer at the front. The results achieved by WRN-28-1-PP are in many cases higher than those of wider WideResNet models without the push-pull layer. In this light, the presence of the push-pull layer is beneficial twofold. It improves the robustness and stability of classification results with respect to noise perturbations of the input images and can favour a reduction of the number of parameters to learn. For instance, the  WRN-28-1-PP has $0.36$M parameters and achieves the same or higher performance that those of bigger models with more parameters (WRN-28-10: $36.4$M parameters; WRN-28-8: $23.3$M parameters; WRN-28-4: $5.8$M parameters). The presence of the push-pull layer can increase the capacity of the networks and their generalization capabilities.

\begin{table*}[t]
\centering
\setlength{\unitlength}{14mm}%
\renewcommand{\arraystretch}{1}
\begin{tabular}{c|cc|C{\unitlength}C{\unitlength}C{\unitlength}C{\unitlength}C{\unitlength}C{\unitlength}C{\unitlength}C{\unitlength}} 
\multicolumn{11}{c}{\bfseries Sensitivity analysis w.r.t. inhibition parameters in  WRN-PP} \\ \hline
~ & ~ & ~ & \multicolumn{8}{c}{noise perturbation ($\sigma^2$)} \\ \cline{4-11}
model & $h$ & $\alpha$ & 0 & 0.0001 & 0.0005 & 0.001 & 0.005 & 0.01 & 0.02 & 0.03 \\ \hline \hline
\multirow{10}{*}{\rotatebox{90}{WRN-16-10}} & - & - & 95.91 & 94.82 & 89.53 & 81.24 & 40.57 & 22.54 & 12.56 & 11.14 \\ \hdashline
~ & 1 & 0.5 & \bfseries 96.01 & \bfseries 95.19 & 90.5 & 82.9 & 42.8 & 27.49 & 17.29 & 13.87 \\
~ & 1 & 1 & 95.83 & 95.14 & 90.42 & 83.38 & 44.05 & 28.83 & 20.55 & 16.94 \\
~ & 1 & 1.5 & 95.81 & 95.03 & 89.89 & 82.15 & 39.21 & 22.68 & 14.54 & 13.07 \\

~ & 1.5 & 0.5 & 95.76 & 95.02 & 91.24 & 85.74 & 54.08 & 37.44 & 26.25 & \bfseries 21.83 \\
~ & 1.5 & 1 &  95.84 & 94.83 & 90.82 & 85.36 & 52.03 & 34.42 & 19.45 & 14.4 \\
~ & 1.5 & 1.5 & 95.67 & 95.09 & 91.68 & 87.18 & 57.59 & 40.18 & 25.37 & 19.53 \\

~ & 2 & 0.5 & 95.62 & 94.72 & 91.46 & 86.82 & 58.85 & \bfseries 41.14 & 24.14 & 18.23 \\
~ & 2 & 1 & 95.45 & 94.03 & 89.9 & 85.06 & 55.44 & 36.09 & \bfseries 24.26 & 20.91 \\
~ & 2 & 1.5 & 95.62 & 95.03 & \bfseries 92.02 & \bfseries 88.02 & \bfseries 58.89 & 36.83 & 20.42 & 15.04 \\ \hline \hline
\end{tabular}
\caption{Sensitivity analysis of the classification accuracy with respect to changes of the configuration parameters of the push-pull layer in a WRN-16-10 model. In bold, we report the best result for each level of Gaussian noise added to the test set.}
\label{tab:sens}
\end{table*}

\subsection{Sensitivity to push-pull parameters}
\label{sec:sensitivity}
We performed an evaluation of the sensitivity of the classification accuracy with respect to variations of the parameters of the push-pull layer, namely the upsampling factor $h$ and the inhibition strenght $\alpha$. In Table~\ref{tab:sens}, we report the results that we achieved with several WRN-16-10-PP models, for which we configured push-pull layers with different parameters. We tested the performance of these models on the CIFAR-10 data set images, which we perturbed with Gaussian noise of different variances. The first row of the tables reports the results of the WRN-16-10 model without the push-pull layer.

From the results Table~\ref{tab:sens} one can notice that no configuration of the parameters of the push-pull layer contributes to achieve the highest classification accuracy on all the noisy versions of the test set. However, using the push-pull layer improves the robustness of the concerned model to noise perturbations of the input images, despite of the specific configuration of its parameters.

It is known from neuro-physiological studies that not all the neurons in area V1 of the visual system of the brain exhibit push-pull inhibition properties~\cite{Taylor595}. In further studies, the upsampling and inhibition strenght parameters of the push-pull layer can be learned from training samples, enforcing sparsity of the inhibition strenght. In this way, only few kernels in the layer will implement inhibition functions, according to what is known to happen in the visual system of the brain.

\section{Conclusions}
We proposed a push-pull layer for CNN architectures, which increases the robustness to noise perturbations of the input images of existing networks and their generalization capabilities. The proposed layer is composed of a set of push and pull convolutions, which implement a non-linear model of an inhibition phenomenon exhibited by some neurons in the visual system of the brain. It can be trained by backpropagation, similarly to convolutional layers.

We validate the effectiveness of the push-pull layer by employing it in state-of-the-art CNN architectures. The results that we achieved using LeNet-5 on the MNIST data set and WideResNet on the CIFAR data sets demontrate that the push-pull layer considerably increases the robustness of existing networks to noise perturbations of the test images.

\section*{Acknowledgements}
This work has been supported by the TrimBot2020 project (H2020 grant no. 688007).

\appendix
\section{Brain-inspired design}
\label{appendix1}
The design of the proposed push-pull block is inspired by neuro-physiological evidence of the presence of a particular form of inhibition, called push-pull inhibition, in the visual system of the brain.

In general, inhibition is the phenomenon of suppression of the response of certain neural receptive fields by means of the action of receptive fields with opposite polarity. From neuro-physiological studies of the visual system of the brain, there is evidence that neurons exhibit various forms of inhibition. 
For instance, end-stopped cells are characterized by an inhibition mechanism that increases their selectivity to line-ending patterns~\cite{Bolz86}. 
In the case of lateral inhibition, the response of a certain neuron suppresses the responses of its neighbouring neurons. Lateral inhibition inspired the design of the Local Response Normalization technique in CNNs, which increased the generalization results of AlexNet~\cite{alexnet}.
Center-surround inhibition is known to increase the detection rate of patterns of interest by suppression of texture in their surroundings, and has been shown to be effective in image processing~\cite{Grigorescu}. 

Neurons that exhibit push-pull inhibition are composed of one receptive field that is excited by a certain positive stimulus (push) and one that is excited by its negative counterpart (pull). In practice, the negative receptive field is larger than the positive one and suppresses its response~\cite{LIU2011542,Li16466}. The effect of push-pull inhibition is to increase the selectivity of neurons for stimulti for which they are tuned, even when they are corrupted by noise~\cite{Freeman2002}.

{\small
\bibliographystyle{ieee}
\bibliography{egbib}
}

\end{document}


\title{Supplementary Material: \\ A Push-Pull Layer Improves Robustness of Convolutional Neural Networks}

\author{Nicola Strisciuglio\\
University of Groningen\\
\\
{\tt\small n.strisciuglio@rug.nl}
\and
Manuel Lopez-Antequera\\
University of Groningen\\
University of Malaga\\
{\tt\small m.lopez.antequera@rug.nl}
\and
Nicolai Petkov\\
University of Groningen\\
\\
{\tt\small n.petkov@rug.nl}
}

\maketitle

\section{Extended results of LeNet and LeNet-PP on the MNIST data set}
In this Section, we report detailed and numerical results achieved by the different configurations of the LeNet and LeNet-PP models on the MNIST data set, as an extension of the results reported in Section 4.1 of the paper. 

\subsection{Gaussian and Poisson noise}
In Tables~\ref{tab:lenet-gaussian} and~\ref{tab:lenet-poisson}, we report the numerical results that correspond to Figures 4 and 5 of the paper, respectively. They refer to experiments on the MNIST data set with Gaussian and Poisson noise.  For details about the configurations of the models, please refer at Table 1 of the paper.

\begin{table*}[h]
\centering
\setlength{\unitlength}{14mm}%
\renewcommand{\arraystretch}{1}
\begin{tabular}{c|C{\unitlength}C{\unitlength}C{\unitlength}C{\unitlength}C{\unitlength}C{\unitlength}C{\unitlength}C{\unitlength}} 
\multicolumn{9}{c}{\bfseries  Results of LeNet and LeNet-PP - MNIST with Gaussian noise} \\ \hline
~ & \multicolumn{8}{c}{noise variance ($\sigma^2$)} \\ \cline{2-9}
model & 0 & 0.001 & 0.01 & 0.1 & 0.2 & 0.3 & 0.4 & 0.5 \\ \hline \hline
A & 98.93 & 98.94 & 98.73 & 94.61 & 86.5 & 76.83 & 68.2 & 61.6 \\ 
PA & \bfseries 99.11 & \bfseries 99.1 & \bfseries 98.93 & \bfseries 95.27 & \bfseries 87.09 & \bfseries 78.38 & \bfseries 69.71 & \bfseries 62.8 \\ \hdashline

B & \bfseries 98.85 & \bfseries 98.88 & 98.64 & 90.24 & 73.91 & 62.62 & 53.55 & 45.67 \\ 
PB & 98.74 & 98.76 & \bfseries 98.66 & \bfseries 95.24 & 87.2 & 78.72 & 71.43 &  63.1 \\ \hdashline

C & 99.06 & 99.04 & 98.84 & 95.02 & 86.2 & 75.79 & 68.18 & 59.08 \\ 
PC & 98.91 & 98.96 & 98.77 & 94.94 & 85.14 & 73.87 & 65.21 & 57.95  \\ \hdashline

D & 98.58 & 98.62 & 98.39 & 90.02 & 78.62 & 68.63 & 60.34 & 53.97 \\ 
PD & 98.84 & 98.77 & 98.57 & 93.41 & 81.99 & 71.27 & 63.88 & 57.24 \\ 
\hline \hline
\end{tabular}
\caption{Results achieved by different LeNet and LeNet-PP models on the MNIST data set with Gaussian noise with increasing variance.}
\label{tab:lenet-gaussian}
\end{table*}
\FloatBarrier

\begin{table*}[h]
\centering
\setlength{\unitlength}{20mm}%
\renewcommand{\arraystretch}{1}
\begin{tabular}{c|C{\unitlength}|C{\unitlength}C{\unitlength}C{\unitlength}C{\unitlength}C{\unitlength}} 
\multicolumn{7}{c}{\bfseries  Results of LeNet and LeNet-PP - MNIST with Poisson noise} \\ \hline
~ & ~ & \multicolumn{5}{c}{contrast ($C$)} \\ \cline{2-7}
model & no noise & 0.5 & 1 & 1.5 & 2 & 2.5 \\ \hline \hline
A & 98.93 & 98.96 & 80.36 & 69.07 & 52.79 & 30.8 \\ 
PA  & 99.11 & 99.11 & 97.53 & 95.09 & 89.1 & 62.84  \\ \hdashline

B & 98.85 & 98.78 & 96.65 & 91.37 & 72.88 & 41.52\\ 
PB  & 98.74 & 98.64 & 97.1 & 94.87 & 91.01 & 75.7\\ \hdashline

C & 99.06 & 98.9 & 97.38 & 92.92 & 79.02 & 45.06 \\ 
PC  & 98.91 & 98.93 & 97.47 & 95.21 & 85.85 & 50.64 \\ \hdashline

D & 98.58 & 98.55 & 79.94 & 63.61 & 46.18 & 37.46 \\ 
PD  & 98.84  & 98.66 & 97.44 & 95.72 & 90.8 & 71.9 \\ 
\hline \hline
\end{tabular}
\caption{Results achieved by different LeNet and LeNet-PP models on the MNIST data set with changes of contrast and Poisson noise.}
\label{tab:lenet-poisson}
\end{table*}
\FloatBarrier

\subsection{Speckle noise}
\begin{figure*}[htb]
\footnotesize
	\begin{center}
		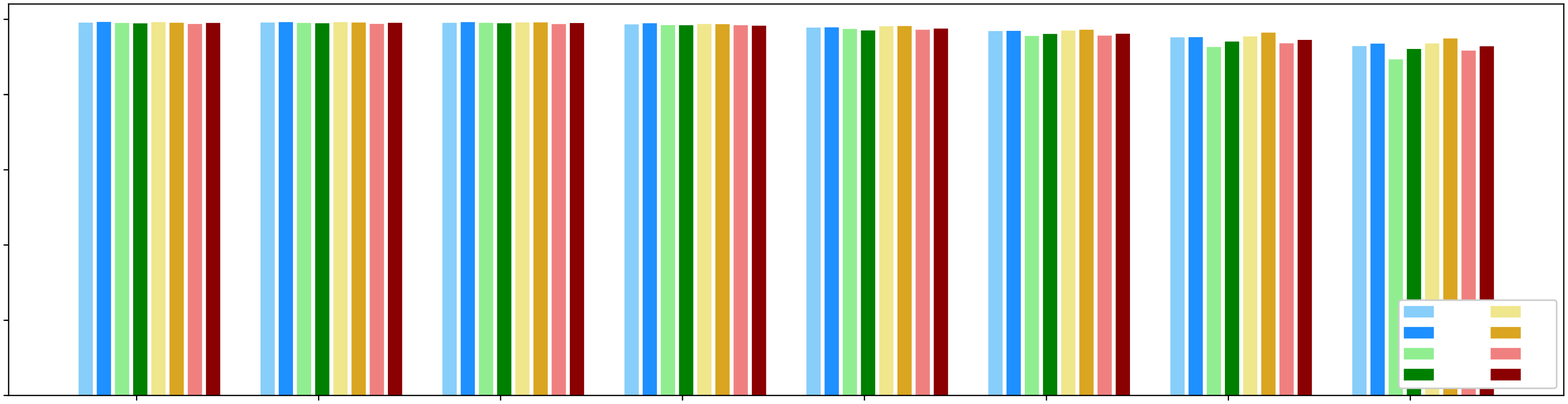
	\end{center}
		\vspace{-6mm}
	\caption{Results of LeNet-5 (lighter colors - A, B, C, D) and push-pull LeNet-5 (darker colors - PA, PB, PC, PD) on the MNIST data set with speckle noise.}
	\label{fig:lenet-speckle}
\end{figure*}

In Figure~\ref{fig:lenet-speckle}, we report results achieved by different configurations of LeNet and LeNet-PP on the MNIST data set corrupted by speckle noise with increasing variance. These results complement those that we reported in Section 4.1 of the paper. In Table~\ref{tab:lenet-speckle}, we report the corresponding numerical results. 

\begin{table*}[h]
\centering
\setlength{\unitlength}{14mm}%
\renewcommand{\arraystretch}{1}
\begin{tabular}{c|C{\unitlength}C{\unitlength}C{\unitlength}C{\unitlength}C{\unitlength}C{\unitlength}C{\unitlength}C{\unitlength}} 
\multicolumn{9}{c}{\bfseries  Results of LeNet and LeNet-PP - MNIST with speckle noise} \\ \hline
~ & \multicolumn{8}{c}{noise variance ($\sigma^2$)} \\ \cline{2-9}
model & 0 & 0.001 & 0.01 & 0.1 & 0.2 & 0.3 & 0.4 & 0.5 \\ \hline \hline
A & 98.93 & 98.96 & 98.88 & 98.47 & 97.62 & 96.69 & 95 & 92.69  \\ 
PA & 99.11 & 99.07 & 99.06 & 98.76 & 97.67 & 96.74 & 95.06 & 93.34  \\ \hdashline

B & 98.85 & 98.85 & 98.87 & 98.27 & 97.26 & 95.41 & 92.44 & 89.15 \\ 
PB & 98.74 & 98.76 & 98.75 & 98.23 & 96.84 & 95.91 & 93.89 & 91.93  \\ \hdashline

C & 99.06 & 99.06 & 98.97 & 98.58 & 97.96 & 96.83 & 95.26 & 93.37 \\ 
PC & 98.91 & 98.95 & 98.99 & 98.52 & 98.03 & 97.05 & 96.28 & 94.72 \\ \hdashline

D & 98.58 & 98.6 & 98.53 & 98.25 & 97.03 & 95.47 & 93.44 & 91.5  \\ 
PD & 98.84 & 98.88 & 98.81 & 98.11 & 97.35 & 95.96 & 94.32 & 92.63 \\ 
\hline \hline
\end{tabular}
\caption{Numerical results achieved on the MNIST data set with speckle noise by different LeNet and LeNet-PP models.}
\label{tab:lenet-speckle}
\end{table*}
\FloatBarrier

\section{Results of WideResNet and WideResNet-PP on the CIFAR data sets}
In this Section, we report detailed and numerical results achieved by the different configurations of the WideResNet and WideResNet-PP models on the CIFAR-10 data set, as an extension of the results reported in Section 4.2 of the paper.

\subsection{Poisson noise}

In Tables~\ref{tab:cifar10-poisson-widen} and~\ref{tab:cifar10-poisson-depth}, we report the numerical results that correspond to Figures 6 and 7 of the paper. We compared the performance of WideResNet models with and without the push-pull layer on the CIFAR-10 data set, corrupted by changes of contrast and addition of Poisson noise to the test images.

\begin{table*}[!h]
\centering
\setlength{\unitlength}{20mm}%
\renewcommand{\arraystretch}{1.1}
\begin{tabular}{c|C{\unitlength}C{\unitlength}C{\unitlength}C{\unitlength}C{\unitlength}} 
\multicolumn{6}{c}{\bfseries Results of WRN and WRN-PP with different widen factors - CIFAR-10 with Poisson noise} \\ \hline
~ & \multicolumn{5}{c}{Contrast ($C$)} \\ \cline{2-6}
model & 0.5 & 1 & 1.5 & 2 & 2.5  \\ \hline \hline
WRN-28-10 & 28.31 & 96.1 & 53.61 & 62.56 & 65.17 \\ 
WRN-28-10-PP & 42.84 & 95.93 & 65.27 & 70.6 & 66.05 \\ \hdashline

WRN-28-8 & 27.52 & 96.07 & 53.22 & 63.04 & 65.33 \\ 
WRN-28-8-PP & 38.84 & 96.07 & 64.38 & 69.09 & 64.85 \\ \hdashline

WRN-28-4 & 29.15 & 95.72 & 49.38 & 60.08 & 62.46 \\ 
WRN-28-4-PP & 36.01 & 95.72 & 58.14 & 64.99 & 64.25 \\ \hdashline

WRN-28-1 & 21.64 & 93.33 & 39.56 & 45.75 & 45.1 \\ 
WRN-28-1-PP & 31.73 & 92.86 & 60.87 & 65.94 & 63.65 \\ 
\hline \hline
\end{tabular}
\caption{Numerical results on CIFAR-10 with Poisson noise, varying the widen factor of the WideResNet architecture. The table refers to Figure 6 of the paper.}
\label{tab:cifar10-poisson-widen}
\end{table*}
 \FloatBarrier

\begin{table*}[!h]
\centering
\setlength{\unitlength}{20mm}%
\renewcommand{\arraystretch}{1.1}
\begin{tabular}{c|C{\unitlength}C{\unitlength}C{\unitlength}C{\unitlength}C{\unitlength}} 
\multicolumn{6}{c}{\bfseries Results of WRN and WRN-PP with different depths - CIFAR-10 with Poisson noise} \\ \hline
~ & \multicolumn{5}{c}{Contrast ($C$)} \\ \cline{2-6}
model & 0.5 & 1 & 1.5 & 2 & 2.5  \\ \hline \hline
WRN-16-10 & 26.22 & 95.91 & 42.57 & 50.67 & 51.4 \\ 
WRN-16-10-PP & 34.85 & 95.45 & 54.77 & 57.39 & 53.19 \\ \hdashline

WRN-22-10 & 28.32 & 95.98 & 52.18 & 62.28 & 63\\ 
WRN-22-10-PP & 36.8 & 95.96 & 62.23 & 68.35 & 66.66 \\ \hdashline

WRN-28-10 & 28.31 & 96.1 & 53.61 & 62.56 & 65.17 \\ 
WRN-28-10-PP & 42.84 & 95.93 & 65.27 & 70.6 & 66.05 \\ \hdashline

WRN-40-10 & 29.98 & 96.08 & 55.82 & 62.3 & 58.85 \\ 
WRN-40-10-PP & 39.9 & 95.91 & 64.48 & 68.77 & 66.68 \\ 
\hline \hline
\end{tabular}
\caption{Numerical results on CIFAR-10 with Poisson noise, varying the depth of the WideResNet architecture. The table refers to Figure 7 of the paper.}
\label{tab:cifar10-poisson-depth}
\end{table*}
\FloatBarrier

\subsection{Gaussian noise}
In Tables~\ref{tab:cifar10-gaussian-widen} and~\ref{tab:cifar10-gaussian-depth}, we report the numerical results that correspond to Figures 8 and 9 of the paper. We compared the performance of WideResNet models with and without the push-pull layer on the CIFAR-10 data set, corrupted by increasing level of Gaussian noise.

\begin{table*}[!h]
\centering
\setlength{\unitlength}{14mm}%
\renewcommand{\arraystretch}{1}
\begin{tabular}{c|C{\unitlength}C{\unitlength}C{\unitlength}C{\unitlength}C{\unitlength}C{\unitlength}C{\unitlength}C{\unitlength}} 
\multicolumn{9}{c}{\bfseries  Results of WRN and WRN-PP with different widen factors - CIFAR-10 with Gaussian noise} \\ \hline
~ & \multicolumn{8}{c}{noise perturbation ($\sigma^2$)} \\ \cline{2-9}
model & 0 & 0.0001 & 0.0005 & 0.001 & 0.005 & 0.01 & 0.02 & 0.03 \\ \hline \hline
WRN-28-10 & 96.1 & 95.33 & 90.94 & 85.52 & 48.56 & 29.68 & 18.58 & 14.55  \\ 
WRN-28-10-PP & 95.93 & 95.33 & 92.38 & 88.94 & 63.15 & 43.61 & 25.86 & 18.32 \\ \hdashline

WRN-28-8 & 96.07 & 95.18 & 91.51 & 85.42 & 47.37 & 28.43 & 18.58 & 15.16 \\ 
WRN-28-8-PP & 96.07 & 95.34 & 92.49 & 88.63 & 60.09 & 37.68 & 21.32 & 16.56\\ \hdashline

WRN-28-4 & 95.72 & 94.91 & 90.87 & 84.86 & 47.79 & 27.11 & 16.89 & 14.16  \\ 
WRN-28-4-PP & 95.72 & 95.02 & 91.32 & 86.55 & 55.91 & 34.41 & 20.14 & 15.16  \\ \hdashline

WRN-28-1 &  93.33 & 92.34 & 86.99 & 79.96 & 40.67 & 22.40 & 12.34 & 10.59  \\ 
WRN-28-1-PP & 92.86 & 92.01 & 88.84 & 83.87 & 50.27 & 30.61 & 18.55 & 14.42  \\ 
\hline \hline
\end{tabular}
\caption{Numerical results on CIFAR-10 with Gaussian noise, varying the widen factor of the WideResNet architecture. The table refers to Figure 8 of the paper.}
\label{tab:cifar10-gaussian-widen}
\end{table*}

\FloatBarrier

\begin{table*}[!h]
\centering
\setlength{\unitlength}{14mm}%
\renewcommand{\arraystretch}{1}
\begin{tabular}{c|C{\unitlength}C{\unitlength}C{\unitlength}C{\unitlength}C{\unitlength}C{\unitlength}C{\unitlength}C{\unitlength}} 
\multicolumn{9}{c}{\bfseries Results of WRN and WRN-PP with different depths - CIFAR-10 with Gaussian noise} \\ \hline
~ & \multicolumn{8}{c}{noise perturbation ($\sigma^2$)} \\ \cline{2-9}
model & 0 & 0.0001 & 0.0005 & 0.001 & 0.005 & 0.01 & 0.02 & 0.03 \\ \hline \hline
WRN-16-10 & 95.91 & 94.82 & 89.53 & 81.24 & 40.57 & 22.54 & 12.56 & 11.14 \\ 
WRN-16-10-PP & 95.45 & 94.03 & 89.9 & 85.06 & 55.44 & 36.09 & 24.26 & 20.91 \\ \hdashline

WRN-22-10 & 95.98 & 95.17 & 91.56 & 85.44 & 46.64 & 27.33 & 17.22 & 13.95\\ 
WRN-22-10-PP & 95.96 & 95.09 & 91.89 & 87.58 & 59 & 38.02 & 22.3 & 17.11  \\ \hdashline

WRN-28-10 & 96.1 & 95.33 & 90.94 & 85.52 & 48.56 & 29.68 & 18.58 & 14.55 \\ 
WRN-28-10-PP & 95.93 & 95.33 & 92.38 & 88.94 & 63.15 & 43.61 & 25.86 & 18.32 \\ \hdashline

WRN-40-10 & 96.08 & 95.29 & 92.07 & 86.82 & 52.98 & 33.77 & 21.39 & 16.58 \\ 
WRN-40-10-PP & 95.91 & 94.74 & 91.66 & 87.95 & 63.35 & 4\bfseries 4.83 & 28.96 & 21.58 \\ 
\hline \hline
\end{tabular}
\caption{Numerical results on CIFAR-10 with Gaussian noise, varying the depth of the WideResNet architecture. The table refers to Figure 9 of the paper.}
\label{tab:cifar10-gaussian-depth}
\end{table*}
\FloatBarrier

\subsection{Speckle noise}
\begin{figure*}[!h]
\footnotesize
	\begin{center}
		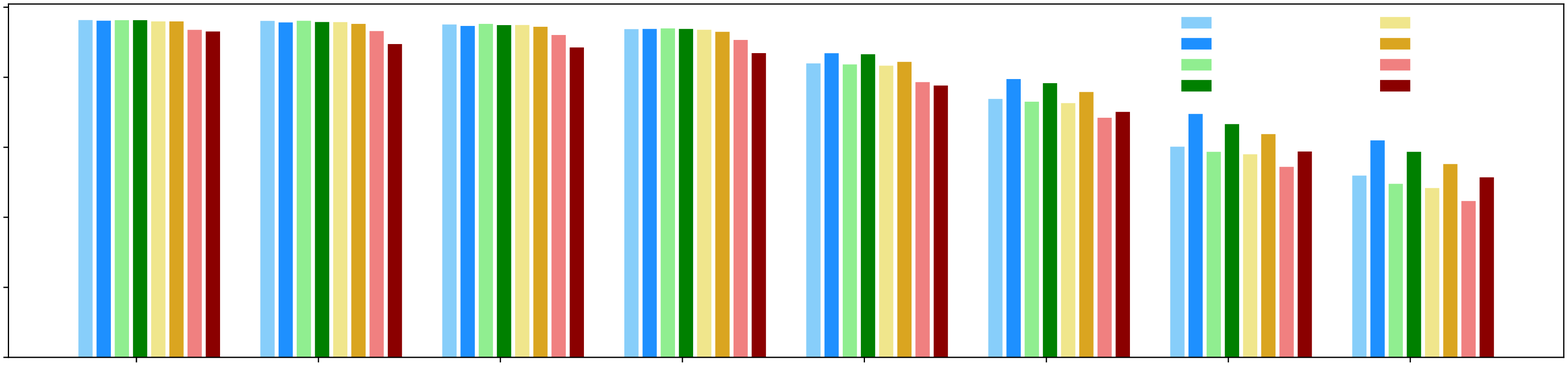
	\end{center}
		\vspace{-6mm}
	\caption{Results achieved on CIFAR-10 with speckle noise by WRN architectures with different widen factors. The Poisson noise is applied after changing the contrast $C$ of the original images.}
	\label{fig:cifar10-speckle-widen}
\end{figure*}

\begin{figure*}[h]
\footnotesize
	\begin{center}
		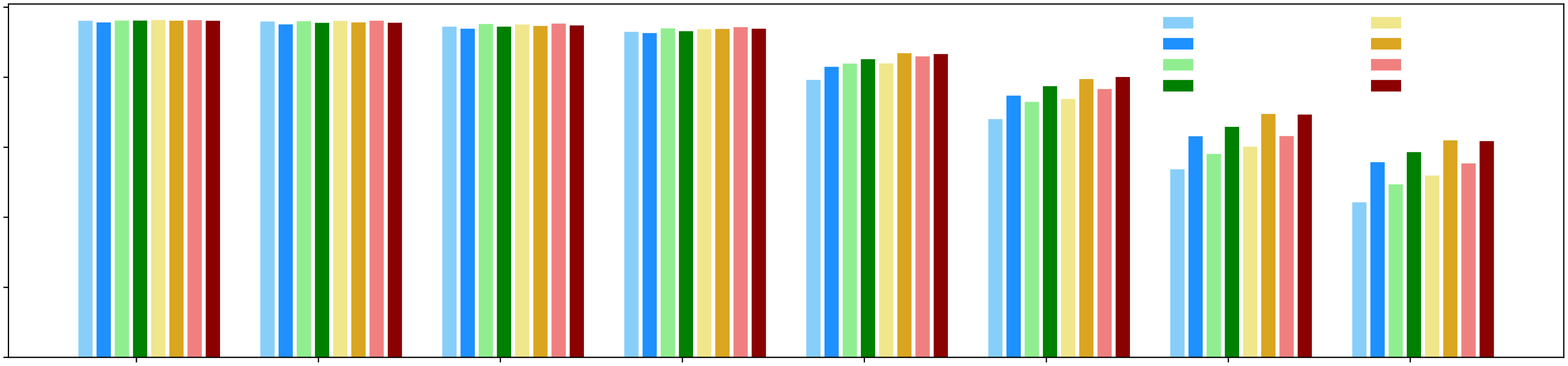
	\end{center}
		\vspace{-6mm}
	\caption{Results achieved on CIFAR-10 with speckle noise by WRN architectures with and without Push-Pull layer with different depths. The Poisson noise is applied after changing the contrast $C$ of the original images.}
	\label{fig:cifar10-speckle-depth}
\end{figure*}

In Figures~\ref{fig:cifar10-speckle-widen} and~\ref{fig:cifar10-speckle-depth} (of this supplementary material paper), we report the numerical results achieved on the CIFAR-10 data set with speckle noise by WideResNet and WideResNet-PP architectures for which we vary the widen factor and the detph, respectively. In Tables~\ref{tab:cifar10-speckle-widen} and~\ref{tab:cifar10-speckle-depth}, we report the corresponding detailed numerical results.

\FloatBarrier

\begin{table*}[!h]
\centering
\setlength{\unitlength}{14mm}%
\renewcommand{\arraystretch}{1.1}
\begin{tabular}{c|C{\unitlength}C{\unitlength}C{\unitlength}C{\unitlength}C{\unitlength}C{\unitlength}C{\unitlength}C{\unitlength}} 
\multicolumn{9}{c}{\bfseries  Results of WRN and WRN-PP with different widen factors - CIFAR-10 with speckle noise} \\ \hline
~ & \multicolumn{8}{c}{noise perturbation ($\sigma^2$)} \\ \cline{2-9}
model & 0 & 0.0001 & 0.0005 & 0.001 & 0.005 & 0.01 & 0.02 & 0.03 \\ \hline \hline
WRN-28-10 & 96.1 & 95.88 & 94.85 & 93.52 & 83.73 & 73.6 & 59.95 & 51.69 \\ 
WRN-28-10-PP & 95.93 & 95.45 & 94.44 & 93.59 & 86.65 & 79.28 & 69.29 & 61.74 \\ \hdashline

WRN-28-8 & 96.07 & 95.9 & 95.03 & 93.74 & 83.43 & 72.8 & 58.49 & 49.34 \\ 
WRN-28-8-PP & 96.07 & 95.56 & 94.67 & 93.59 & 86.35 & 78.08 & 66.4 & 58.48\\ \hdashline

WRN-28-4 & 95.72 & 95.54 & 94.69 & 93.35 & 83.08 & 72.38 & 57.77 & 48.14 \\ 
WRN-28-4-PP & 95.72 & 95.03 & 94.19 & 92.75 & 84.19 & 75.56 & 63.55 & 54.99  \\ \hdashline

WRN-28-1 & 93.33 & 92.96 & 91.85 & 90.44 & 78.4 & 68.22 & 54.18 & 44.43  \\ 
WRN-28-1-PP & 92.86 & 89.25 & 88.29 & 86.67 & 77.42 & 69.89 & 58.56 & 51.2 \\ 
\hline \hline
\end{tabular}
\caption{Numerical results on CIFAR-10 with speckle noise, varying the widen factor of the WideResNet architecture.}
\label{tab:cifar10-speckle-widen}
\end{table*}
\FloatBarrier

\FloatBarrier

\begin{table*}[h]
\centering
\setlength{\unitlength}{14mm}%
\renewcommand{\arraystretch}{1.1}
\begin{tabular}{c|C{\unitlength}C{\unitlength}C{\unitlength}C{\unitlength}C{\unitlength}C{\unitlength}C{\unitlength}C{\unitlength}} 
\multicolumn{9}{c}{\bfseries Results of WRN and WRN-PP with different depths - CIFAR-10 with speckle noise} \\ \hline
~ & \multicolumn{8}{c}{noise perturbation ($\sigma^2$)} \\ \cline{2-9}
model & 0 & 0.0001 & 0.0005 & 0.001 & 0.005 & 0.01 & 0.02 & 0.03 \\ \hline \hline
WRN-16-10 & 95.91 & 95.69 & 94.23 & 92.77 & 79.02 & 67.84 & 53.5 & 44.05 \\ 
WRN-16-10-PP & 95.45 & 94.89 & 93.64 & 92.4 & 82.78 & 74.55 & 62.93 & 55.51 \\ \hdashline

WRN-22-10 & 95.98 & 95.79 & 95 & 93.72 & 83.67 & 72.73 & 57.88 & 49.19\\ 
WRN-22-10-PP & 95.96 & 95.32 & 94.24 & 92.95 & 84.94 & 77.25 & 65.64 & 58.4  \\ \hdashline

WRN-28-10 & 96.1 & 95.88 & 94.85 & 93.52 & 83.73 & 73.6 & 59.95 & 51.69 \\ 
WRN-28-10-PP & 95.93 & 95.45 & 94.44 & 93.59 & 86.65 & 79.28 & 69.29 & 61.74 \\ \hdashline

WRN-40-10 & 96.08 & 95.93 & 95.11 & 94.07 & 85.74 & 76.42 & 62.97 & 55.15 \\ 
WRN-40-10-PP & 95.91 & 95.35 & 94.58 & 93.63 & 86.42 & 79.86 & 69.14 & 61.54 \\ 
\hline \hline
\end{tabular}
\caption{Numerical results on CIFAR-10 with speckle noise, varying the depth of the WideResNet architecture.}
\label{tab:cifar10-speckle-depth}
\end{table*}
\FloatBarrier